\documentclass{article}
\PassOptionsToPackage{numbers, compress}{natbib}

\usepackage[dblblindworkshop, final]{neurips_2026}

\usepackage[utf8]{inputenc} 
\usepackage[T1]{fontenc}    
\usepackage{hyperref}       
\usepackage{url}            
\usepackage{booktabs}       
\usepackage{amsfonts}       
\usepackage{nicefrac}       
\usepackage{microtype}      
\usepackage{xcolor}         
\usepackage[pdftex]{graphicx}
\usepackage{adjustbox}

\usepackage[table]{xcolor}
\usepackage[x11names]{xcolor}

\title{LEAF: LUT-aware Efficient Attention on FPGAs}

\author{ 
  \bfseries Arianna Cox$^{1}$,
  Chang Sun$^{2}$,
  Lauri Laatu$^{1}$,
  Benedikt Maier$^{1}$,
  Aadeesh Sharma$^{1}$,\\
  \bfseries Leo Rozanov$^{1}$,
  Wayne Luk$^{1}$,
  Maria Spiropulu$^{2}$,
  Alexander Tapper$^{1}$,
  Zhiqiang Que$^{1, 3}$
  \\[1.0em]
  \begin{tabular}{c}
    $^{1}$Imperial College London, United Kingdom \\
    $^{2}$California Institute of Technology, USA \\
    $^{3}$University of Bristol, United Kingdom
  \end{tabular}
}

\begin{document}

\maketitle

\begin{abstract}

  We introduce Lookup-table (LUT)-aware Efficient Attention on FPGAs (LEAF). LEAF is the first nanosecond-scale LUT-aware transformer for FPGAs. Deploying transformers that are large enough to perform well, small enough to fit on an FPGA, and fast enough to meet the sub-microsecond timescales required by real-time systems such as those at CERN remains difficult, since the same attention mechanism that gives these models their high expressivity also makes them costly to implement in hardware.
  LEAF converts the query, key, and value projections and the final output feed-forward layer into FPGA-native layers with LUT-aware training, creating the first LUT-aware transformer and linformer layers built for ultra-low-latency inference.
  Attention resists LUT-aware training because it multiplies two activations where the method needs a constant weight to tabulate; LEAF tabulates the attention block's constant matrix-vector multiplications and leaves the rest in arithmetic form.
  On a standard jet tagging benchmark this improves accuracy by 1.8 percentage points while using 2.9 times fewer LUTs than the previous state-of-the-art sub-microsecond transformer at the same input size. Our most efficient LEAF-based linformer with 64 input particles reaches 80.4\% classification accuracy at 28\,k LUTs and 35\,ns.

\end{abstract}

\section{Introduction}

The CMS~\cite{CMS:2008xjf} and ATLAS~\cite{atlas} experiments at the Large Hadron Collider~\cite{lhc1995large} at CERN generate hundreds of terabytes of data per second from 40~MHz collisions. Since this data cannot all feasibly be stored, the experiments select a small subset of potentially interesting events using a two-level trigger system, consisting of an FPGA-based hardware trigger followed by a software trigger. The experiments can only store the full detector readout for a few microseconds~\cite{cms-tdr-021,atlas-tdr-029}, due to limited buffer capacity, and thus the hardware trigger must decide whether to keep each event within this time frame. Any events not selected are permanently discarded, making trigger efficiency crucial for the experiments' physics reach. To improve filtering performance on the hardware trigger, physicists aim to deploy advanced ML algorithms, which demand highly compressed models capable of running on FPGAs within the resource and latency budgets.

Jet tagging, the task of identifying which particle initiated a collimated spray of particles, is a key step in these trigger systems: accurately tagged jets provide an essential handle for distinguishing rare signals of interest from overwhelming Standard Model backgrounds. Jet tagging has also become a standard benchmark within the fast machine learning community, widely used to demonstrate quantization and compression techniques for FPGA-based neural networks, and we therefore use it to compare our LEAF workflow against previous methods.

LUT-aware training replaces arithmetic with truth tables learned during training, absorbing each constant weight into a table indexed by a single input. Attention resists this: the operation that defines it multiplies two activations and offers no constant to absorb, so existing LUT-aware methods~\cite{hgqlut} provide dense and convolutional layers only. Within the attention block, however, the query, key and value projections and the output feed-forward layer are constant matrix-vector multiplications (CMVMs). We convert all of them to table-based representations, while the activation-activation products remain in arithmetic form. All parts are trained under a single resource budget, so the trade-off between accuracy and hardware cost is optimized across the whole network. We call the resulting LUT-aware transformer and linformer layers LEAF. Combined with HGQ-LUT~\cite{hgqlut} for training and Alkaid~\cite{alkaid} for bit-exact RTL generation, they form an end-to-end workflow that, on the hls4ml jet tagging benchmark, improves accuracy by 1.8 percentage points while using 2.9$\times$ fewer LUTs than the previous state-of-the-art sub-microsecond transformer~\cite{firstTransformer} with 32 input particles. We have integrated our transformer and linformer layers into the open-source HGQ~\cite{hgq} repository, enabling the fast machine learning community to develop efficient LUT-aware transformers for their own applications.

\section{LUT-Aware Transformer Workflow}
\label{section:workflow}

\subsection{LUT-Aware Training using HGQ-LUT}

High Granularity Quantization (HGQ)~\cite{hgq} is a quantization-aware training method that assigns each weight, bias and activation its own trainable bitwidth, rather than applying a single precision uniformly across a layer or network. It optimizes these bitwidths jointly with the network weights by adding a differentiable resource surrogate, Effective Bit Operations (EBOPs), to the loss function. This allows standard gradient descent to trade off accuracy against estimated hardware cost automatically, including automatic pruning when bitwidths reach zero. This fine-grained approach yields highly compact, hardware-efficient models suited to ultra-low-latency FPGA inference, without requiring manual bitwidth tuning.

HGQ-LUT~\cite{hgqlut} extends the HGQ method to LUT-based neural networks, a class of architectures that map neural network operations directly onto logic LUTs (L-LUTs) rather than mapping to standard multiply-accumulate (MAC) operations. HGQ-LUT neural network layers are built from single-input logical LUTs, each implemented during training as a small multi-layer perceptron (MLP) and compiled into L-LUT truth tables for FPGA deployment. HGQ-LUT applies HGQ's heterogeneous differentiable quantization to the inputs and outputs of each L-LUT, extending the EBOPs resource surrogate to this LUT-based setting so that the trade-off between model accuracy and LUT usage is explored during gradient descent, including automatic L-LUT pruning.

Our LEAF attention replaces all CMVM operations, specifically in the query, key, and value projections and the output feed-forward layer, with table-based operations and LUT-aware layers. The key-query multiplication and subsequent multiplication by the value matrix, in contrast, involve variable-variable multiplication and therefore remain in conventional arithmetic form. However, their resource usages contribute to the same differentiable EBOPs resource surrogate as the LUT-aware layers, so that the trade-off between accuracy, arithmetic structure and resource usage is optimized consistently across the entire network. Our LEAF transformer and linformer layers are available in the open-source HGQ tool.

\subsection{RTL Generation using Alkaid}

For streamlined deployment of our LEAF models, we used Alkaid~\cite{alkaid}, a compiler framework that can translate pretrained Python neural networks directly into RTL, while also applying further optimizations to compress the design. Through Alkaid's built-in HGQ plugin, our training-to-RTL workflow remains simple and bit-exact verifiable, combining the HGQ tool for model training with Alkaid's efficient hardware generation.

\section{Application to Jet Tagging}

\subsection{Jet Tagging Task}

We demonstrate the performance of our LEAF transformer workflow on a jet tagging task using the HLS4ML LHC Jet dataset~\cite{hls4ml-dataset}. The dataset contains jets of up to 150 particles, each initiated by a gluon, light quark, W boson, Z boson, or top quark, giving 5 classes in total. This dataset has been used extensively to benchmark FPGA-based jet taggers in prior works~\cite{firstTransformer, mlpm-fpga, deepsetqkeras, nas-fpga}, and we compare our results directly against these in Section~\ref{section:results}.

\subsection{Model Details}

The model architecture used in this work is an encoder-only transformer with residual connections using a standard attention layer~\cite{transformer} with a single head. The model input is a sequence of particles, each represented by three features. The three particle features are transverse momentum $p_\mathrm{T}$, pseudorapidity $\eta$, and azimuthal angle $\phi$. Particles are ordered by decreasing $p_\mathrm{T}$, and separate models are trained using only $n$ of the highest-$p_\mathrm{T}$ particles, with $n$ being 8, 16, 32 or 64. The model outputs a probability distribution over the five jet classes. Model accuracy is defined as the percentage of jets for which the highest-probability class is correct.

We also investigated linformer architectures~\cite{linformer} since the projection of the key and value matrices to lower dimensions means that the computational complexity of the attention layer scales linearly with input sequence length rather than quadratically for transformers. The linformer attention layers are by design less resource intensive, so they can more easily be compressed to the required size to fit on the target FPGA.

Our LEAF models are trained using the HGQ tool and are optimized and converted to RTL using Alkaid, as set out in Section~\ref{section:workflow}. Note that in our work, we apply saturation (SAT) mode rather than wrap-around (WRAP) mode to the input and output quantizers of each L-LUT. This RTL is then synthesized using Vivado for a Xilinx XCU 250 FPGA to produce the hardware results laid out in Section~\ref{section:results}. Our code\footnote{https://github.com/arianna-cox/LEAF\_transformer} is available both for reproducibility of the results in this paper and to act as a useful example for others seeking to develop LEAF transformers and linformers for their own applications.

\begin{figure}[b]
  \centering
  \includegraphics[width=0.95\linewidth]{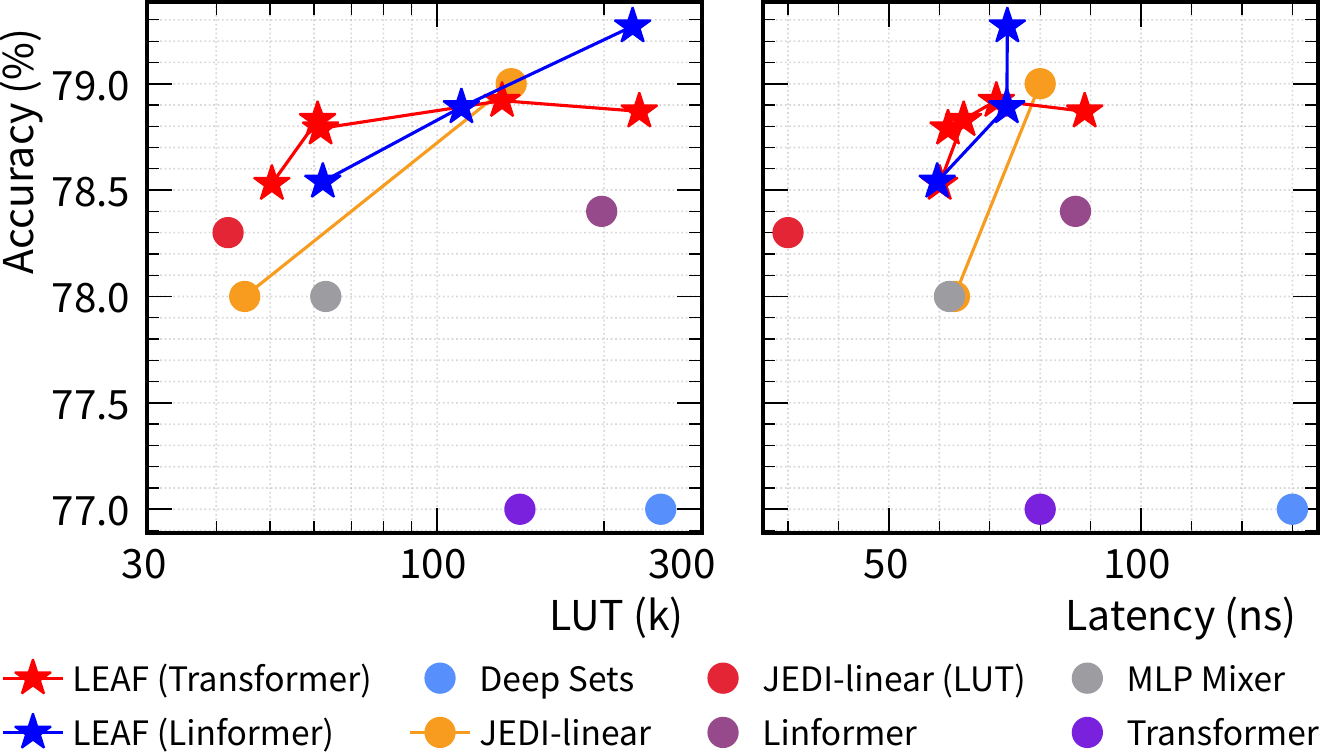}
  \caption{Accuracy against LUT usage and latency for models with 32 input particles.}
  \label{fig:ParetoFront}
  \vspace{-0.5cm}
\end{figure}

\subsection{Results}
\label{section:results}

We compare our LEAF models against prior works across five key metrics: accuracy, latency, LUT usage, DSP usage, and initiation interval, evaluating each for 8, 16, 32 or 64 input particles. We evaluate our model performance against the prior state-of-the-art sub-microsecond transformer and linformer models~\cite{firstTransformer, alkaid}, as well as several alternatives, including JEDI-linear~\cite{jedi-linear, hgqlut}, MLP Mixer~\cite{mlpm-fpga}, and Deep Sets~\cite{deepsetqkeras}. The full jet tagging comparison is presented in Table~\ref{tab:comparison}, with accuracy shown against LUT usage and latency for all 32-input particle models in Figure~\ref{fig:ParetoFront}.

These results show that LEAF's LUT-aware attention with HGQ-LUT is a major improvement on the HGQ method alone for transformer architectures~\cite{firstTransformer}. To isolate the contribution of LUT-aware attention, we separate LEAF’s gains over prior work into the part due to the Alkaid backend and the part due to LEAF itself. In Table~\ref{tab:comparison}, we compare the resource usage and latency between HGQ transformer models converted with da4ml~\cite{firstTransformer} against the same HGQ models converted with Alkaid~\cite{alkaid}. For example, for the 32-particle transformer model, LUTs go from 180k with da4ml to 141k with Alkaid. However, when using LEAF with Alkaid we obtain both a 1.8 percentage point increase in accuracy and a reduction to 62k LUT. In this case roughly a third of the saving comes from the backend and two-thirds from the LUT-aware training. This comparison shows that the majority of the improvement achieved by our LEAF transformers and linformers is attributable to LUT-aware training and table-based mapping, although the switch from da4ml to Alkaid also contributes, through additional optimizations such as affine interval tracking and improved distributed arithmetic optimization~\cite{alkaid}.

These results demonstrate that LEAF moves transformers competitively onto the accuracy-resource Pareto front for jet tagging on FPGAs, alongside the leading architecture JEDI-linear~\cite{jedi-linear}.

\begin{table}
    \centering
    \scalebox{0.875}{\begin{tabular}{lcccccccc}
            \toprule
            Architecture Type                   & Datapath & Backend & Particles & Acc. (\%)     & Lat. (ns) & LUT (k) & II (clk) & DSP \\
            \midrule

            \textbf{Transformer (ours)}         & L-LUT    & alkaid  & 8         & 66.3          & 71        & 47      & 1        & 51  \\
            \textbf{Transformer (ours)}         & L-LUT    & alkaid  & 16        & 72.6          & 66        & 66      & 1        & 0   \\
            \textbf{Transformer (ours)}         & L-LUT    & alkaid  & 32        & 78.8          & 62        & 62      & 1        & 0   \\
            \textbf{Transformer (ours)}         & L-LUT    & alkaid  & 64        & 80.9          & 74        & 162     & 1        & 0   \\
            \midrule
            Transformer~\cite{alkaid}           & MAC      & alkaid  & 16        & 72.3          & 98        & 229     & 1        & 0   \\
            Transformer~\cite{alkaid}           & MAC      & alkaid  & 32        & 77.0          & 80        & 141     & 1        & 0   \\
            \midrule
            Transformer~\cite{firstTransformer} & MAC      & da4ml   & 8         & 66.3          & 104       & 246     & 1        & 0   \\
            Transformer~\cite{firstTransformer} & MAC      & da4ml   & 16        & 72.3          & 98        & 279     & 1        & 0   \\
            Transformer~\cite{firstTransformer} & MAC      & da4ml   & 32        & 77.0          & 83        & 180     & 1        & 0   \\
            Transformer~\cite{firstTransformer} & MAC      & da4ml   & 64        & 77.9          & 44        & 47      & 1        & 0   \\ 
            \midrule
            \textbf{Linformer (ours)}           & L-LUT    & alkaid  & 8         & 66.2          & 61        & 50      & 1        & 0   \\
            \textbf{Linformer (ours)}           & L-LUT    & alkaid  & 16        & 73.1          & 68        & 73      & 1        & 5   \\
            \textbf{Linformer (ours)}           & L-LUT    & alkaid  & 32        & 79.3          & 73        & 225     & 1        & 0   \\
            \textbf{Linformer (ours)}           & L-LUT    & alkaid  & 32        & 78.9          & 73        & 111     & 1        & 0   \\
            \textbf{Linformer (ours)}           & L-LUT    & alkaid  & 64        & 80.8          & 57        & 76      & 1        & 0   \\
            \textbf{Linformer (ours)}           & L-LUT    & alkaid  & 64        & 80.4          & 35        & 28      & 1        & 0   \\
            \midrule
            Linformer~\cite{alkaid}             & MAC      & alkaid  & 16        & 72.8          & 102       & 185     & 1        & 0   \\
            Linformer~\cite{alkaid}             & MAC      & alkaid  & 32        & 78.4          & 87        & 198     & 1        & 0   \\
            \midrule
            Linformer~\cite{firstTransformer}   & MAC      & da4ml   & 8         & 66.3          & 110       & 230     & 1        & 0   \\
            Linformer~\cite{firstTransformer}   & MAC      & da4ml   & 16        & 72.8          & 105       & 246     & 1        & 0   \\
            Linformer~\cite{firstTransformer}   & MAC      & da4ml   & 32        & 78.4          & 140       & 267     & 1        & 0   \\
            Linformer~\cite{firstTransformer}   & MAC      & da4ml   & 64        & 79.8          & 78        & 202     & 1        & 0   \\
            \midrule
            JEDI-linear~\cite{hgqlut}           & L-LUT    & da4ml   & 32        & 78.3          & 30        & 42      & 1        & 0   \\
            \midrule
            JEDI-linear~\cite{jedi-linear}      & MAC      & da4ml   & 8         & 66.5          & 79        & 136     & 1        & 0   \\
            JEDI-linear~\cite{jedi-linear}      & MAC      & da4ml   & 16        & 73.6          & 75        & 136     & 1        & 0   \\
            JEDI-linear~\cite{jedi-linear}      & MAC      & da4ml   & 32        & 79.0          & 80        & 136     & 1        & 0   \\
            JEDI-linear~\cite{jedi-linear}      & MAC      & da4ml   & 64        & 81.8          & 78        & 164     & 1        & 0   \\
            \midrule
            MLP Mixer~\cite{mlpm-fpga}          & MAC      & da4ml   & 16        & 71.7          & 68        & 75      & 1        & 0   \\
            MLP Mixer~\cite{mlpm-fpga}          & MAC      & da4ml   & 32        & 78.0          & 62        & 63      & 1        & 0   \\
            MLP Mixer~\cite{mlpm-fpga}          & MAC      & da4ml   & 64        & 79.7          & 72        & 159     & 1        & 0   \\
            \midrule
            Deep Sets~\cite{deepsetqkeras}$^1$  & MAC      & hls4ml  & 8         & $\le$64.0$^2$ & 95        & 386     & 3        & 626 \\
            Deep Sets~\cite{deepsetqkeras}$^1$  & MAC      & hls4ml  & 16        & $\le$69.4$^2$ & 115       & 747     & 3        & 555 \\
            Deep Sets~\cite{deepsetqkeras}$^1$  & MAC      & hls4ml  & 32        & $\le$75.9$^2$ & 130       & 903     & 2        & 434 \\
            \bottomrule
        \end{tabular}}
    \vspace{0.2cm}
    \caption{
        Performance comparison of models on the hls4ml LHC Jet dataset~\cite{hls4ml-dataset}. All results presented are from out-of-context results post place \& route on Vivado 2025.1 if not otherwise specified.\\
        \footnotesize{
            $^1$: The Deep Sets models are from logic synthesis results in~\cite{deepsetqkeras}. The authors did not report post place or routing numbers. Vivado and Vitis 2020.1 are used for the synthesis of these models~\cite{deepsetqkeras}.\\
            $^2$: The accuracy of the Deep Sets models after quantization and HLS conversion is not specified. The quoted accuracy is the accuracy of the original floating-point models, where the authors mention that the accuracy of the deployed models are at least $90$\% of the floating point accuracy.\\
            }
    }
    \vspace{-1.5cm}
    \label{tab:comparison}
\end{table}

\section{Conclusions}

We have presented LEAF, the first nanosecond-scale LUT-aware transformer for FPGA applications, achieving greater model accuracy with fewer LUTs than previous transformer implementations. By contributing support for our transformer and linformer layers in the open-source HGQ repository, we aim to make efficient, LUT-aware attention mechanisms broadly accessible to the fast machine learning community. We envision that these LUT-aware transformers could prove useful and effective across a wide range of high-energy physics tasks beyond jet tagging, including anomaly detection and foundation models. We believe our workflow makes larger, more expressive transformers more easily deployable for real-time systems, from CERN trigger systems to other ultra-low-latency applications.

\section*{Acknowledgments}

Work done by Imperial College is funded by the Science and Technology Facilities Council (STFC) grant ST/W000636/1
and Intel and Lenovo (ICICLE HPC \& AI partnership).
Work done at Caltech is partially funded by the U.S. Department of Energy grant numbers DE-SC0011925, DE-FOA-0002705 and National Science Foundation grant numbers PHY240298, PHY2117997.

\bibliographystyle{unsrturl}
\bibliography{bibliography}

\end{document}